# An Ontology for Comprehensive Tutoring of Euphonic Conjunctions of Sanskrit Grammar

**Rajitha V.**
*Department of Computer Science, Meenakshi College for Women, Chennai, India*
*Research Scholar, Mother Teresa Women's University, Kodaikanal, India*

**Kasmir Raja S. V.**
*Dean – Research, SRM University, Chennai, India*

**Meenakshi Lakshmanan**
*Department of Computer Science, Meenakshi College for Women, Chennai, India*
E-Mail: meenakshi.lakshmanan@gmail.com

## Abstract

Euphonic conjunctions (*sandhis*) form a very important aspect of Sanskrit morphology and phonology. The traditional and modern methods of studying about euphonic conjunctions in Sanskrit follow different methodologies. The former involves a rigorous study of the Pāṇinian system embodied in Pāṇini's *Āṣṭādhyāyī*, while the latter usually involves the study of a few important *sandhi* rules with the use of examples. The former is not suitable for beginners, and the latter, not sufficient to gain a comprehensive understanding of the operation of *sandhi* rules. This is so since there are not only numerous *sandhi* rules and exceptions, but also complex precedence rules involved. The need for a new ontology for *sandhi*-tutoring was hence felt. This work presents a comprehensive ontology designed to enable a student-user to learn in stages all about euphonic conjunctions and the relevant aphorisms of Sanskrit grammar and to test and evaluate the progress of the student-user. The ontology forms the basis of a multimedia sandhi tutor that was given to different categories of users including Sanskrit scholars for extensive and rigorous testing.

**Keywords:** Sanskrit, euphonic conjunction, sandhi, Panini, sandhi tutor, ontology

## 1. Introduction

The phenomenon of euphonic conjunctions is prevalent in a significant way in many Indian languages. In Sanskrit this phenomenon has far-reaching effects at both the morphological and phonological levels, and has also been dealt with in a most precise and comprehensive manner in the language's grammar specification. A euphonic conjunction ('*sandhi*' in Sanskrit) causes a word to undergo changes either due to internal factors or due to the influence of neighboring words.

Sanskrit grammar is codified into a precise formulation contained in the work called the '*Aṣṭādhyāyī*', which is widely acknowledged as the earliest known work on descriptive linguistics and the final authority on Sanskrit grammar. This magnum opus of the ancient grammarian Pāṇini, contains tersely stated aphorisms or '*sūtras*' that lay out the grammatical rules of the language in a way that proves that Sanskrit is a language that is both natural and formal.

Traditional teaching-learning methods of Sanskrit grammar involve the complete study of the *Aṣṭādhyāyī*, including learning all the nearly 4000 *sūtras* by rote. Further, one or more commentaries on the *Aṣṭādhyāyī* are imperative to get a complete understanding of the *sūtras*. *Sandhi* learning would automatically form part of this grand exercise that takes years to complete. On the other hand, existing books, tools or websites that tutor beginners on Sanskrit grammar or even in euphonic conjunctions in particular, do so only in a general way [6-14]. Though the tutoring method followed is based on general categorizations of *sandhis* such as vowel-based and consonant-based *sandhis*, a grasp of the Pāṇinian system or of subtle nuances with regard to the application of *sandhi* rules is not had from them, making this method less comprehensive than would be acceptable to a Sanskrit scholar. For instance, the ordering of rules is of critical importance in the Pāṇinian system, and a knowledge of this is required to comprehensively grasp how *sandhis* operate. There is thus a recognized need for a conceptualization of a tutoring scheme and a systematic specification of the same. The comprehensive *sandhi*-tutoring ontology presented in this work fulfils this need to cater to a beginner's requirements as well as to make sure that such a learner gains complete knowledge of the operation of euphonic conjunctions in Sanskrit, along with the relevant Pāṇinian aphorisms.

A computational algorithm for generating euphonic conjunctions was earlier developed by the authors [3-4]. With this methodology for processing search-related euphonic conjunctions as the basis, an ontology for teaching euphonic conjunctions has been developed for the first time in this work. This unique ontology forms the basis of a multimedia *sandhi* tutor that has been developed to enable a student-user to learn step-by-step all about euphonic conjunctions and the relevant Pāṇinian aphorisms, tests and evaluates the progress of the student-user, and when words or word-components are given as input, generates all the pertinent, permissible euphonic conjunctions and also spells out grammatical rules that apply. The tutor includes a voice component because *sandhi* can cause phonological changes as well. The ontology has been tested and validated through the rigorous testing and validation of the *sandhi* tutoring tool by Sanskrit scholars. A survey of users of two other categories has also been taken and their feedback analyzed.

## 2. Some Language Preliminaries

### 2.1. The Sanskrit alphabet

The Sanskrit alphabet are broadly classified into vowels (*svaras*) and consonants (*vyañjanas*). These are further categorized as depicted below:

**Vowels**

1. 5 short vowels or *hrasva* (*a, i, u, ṛ, ḷ*) – these take one unit of time to pronounce
2. 8 long vowels or *dīrgha* (*ā, ī, ū, ṝ, e, ai, o, au*) – these take two units of time to pronounce
3. 8 vowels with three units of pronunciation time or *pluta* (denoted as *ā3, ī3*, etc.)
4. 1 *avagraha* (denoted by a single apostrophe in Latin E-text and as 'ऽ' in *Devanāgarī* script and not pronounced or pronounced as '*a*' in half a unit of time.
   In general, the first two categories are counted to give 13 vowels.

**Consonants**

1. 4 semi-vowels or *antasthās* (*y, r, l, v*)
2. 25 mutes or *sparśas*
      i. 20 non-nasal mutes (*k, kh, g, gh, c, ch, j, jh, ṭ, ṭh, ḍ, ḍh, t, th, d, dh, p, ph, b, bh*)
      ii. 5 nasals or *anunāsikas* (*ṅ, ñ, ṇ, n, m*)
      These 25 mutes are otherwise categorized as:
         a. Gutturals (*k, kh, g, gh, ṅ*)

   b. Palatals (*c, ch, j, jh, ñ*)
   c. Cerebrals (*ṭ, ṭh, ḍ, ḍh, ṇ*)
   d. Dentals (*t, th, d, dh, n*)
   e. Labials (*p, ph, b, bh, m*)
3. 3 sibilants (*ś, ṣ, s*)
4. 1 aspirate (*h*)
   (The sibilants and aspirate are together called *ūṣmāṇas*)
   The above four categories make up 33 consonants.
5. 1 *anusvāra* (*ṁ*)
6. 1 *visarga* (*ḥ*)
7. 1 *jihvāmūlīya* (*visarga* pronounced as at the end of '*kaḥ*')
8. 1 *upadhmānīya* (*visarga* pronounced as at the end of '*paf*')
   Thus there are 4 special consonants.

In all, there are 50 letters in the Sanskrit alphabet apart from the 8 *pluta* vowels and the *avagraha*.

The 13 vowels and 33 consonants are laid out in the *Māheśvara-sūtras*, aphorisms that list these letters in a non-trivial order. The interpretation of Pāṇini's aphorisms are heavily dependent on the ordering presented in the total of 14 *Māheśvara-sūtras*, which are presented below.

1. *a-i-u-ṇ*
2. *ṛ-ḷ-k*
3. *e-o-ṅ*
4. *ai-au-c*
5. *ha-ya-va-ra-ṭ*
6. *la-ṇ*
7. *ña-ma-ṅa-ṇa-na-m*
8. *jha-bha-ñ*
9. *gha-ḍha-dha-ṣ*
10. *ja-ba-ga-ḍa-da-ś*
11. *kha-pha-cha-ṭha-tha-ca-ṭa-ta-v*
12. *ka-pa-y*
13. *śa-ṣa-sa-r*
14. *ha-l*

The last letter in each of these aphorisms is only a place-holder. The first four aphorisms list only the short forms of all the vowels, while the rest list the semi-vowels and consonants; the latter list has the vowel '*a*' appended to each letter only to enable pronunciation of the semi-vowel or consonant.

## 2.2. Types of euphonic conjunctions

Euphonic conjunctions in Sanskrit cause different types of transformations to occur. Let $x$ and $y$ be the adjacent letters that can potentially cause such a transformation. Let $u$ and $w$ respectively denote the letter preceding $x$ and that succeeding $y$. The possible types and sub-types of transformations are listed below.

1. Substitution (*ādeśaḥ*)
   i. $x$ is replaced
      Example: *viṣṇo iha = viṣṇav iha*
   ii. $y$ is replaced
      Example: *hare ava = hare 'va*
   iii. Both $x$ and $y$ are replaced
      Example: *ramā iva = rameva*
2. Deletion (*lopaḥ*)
   i. $x$ or $y$ is deleted
      Example: *viṣṇav iha = viṣṇa iha*

3. Addition (*āgamaḥ*)
   i. A new letter *z* is added between *x* and *y*
      Example: *rāma chatram = rama**t** chatram*
   ii. *x* or *y* is duplicated
      Example: *sa**n** acyutaḥ = sa**nn** acyutaḥ*
4. No transformation (*prakṛtibhāvaḥ*)
   i. No change is made
      Example: *harī etau = harī etau* (if the word *'harī'* is in dual number)

## 3. Challenges in Designing an Effective Sandhi-tutoring Ontology

For this work, a thorough study of the original *Aṣṭādhyāyī* was undertaken with the help of the authoritative commentaries, *Siddhānta-kaumudī* [1] and *Kāśikā* [2]. Through this extensive study and analysis, more than 100 *sūtras* of Pāṇini were identified as pertaining specifically to *sandhis* or used in their understanding. A unique ontology was then developed for *sandhi*-tutoring based on this study and on the implementation of a *sandhi* processor developed by the authors in [4].

Pāṇini's *Āṣṭādhyāyi* comprises eight chapters, each of which is divided into four quarters or *pādas*. The entire work is considered as comprising two parts. Part 1 consists of the first seven chapters and the first quarter of the eighth chapter, and is called the '*sapādasaptādhyāyī*'. Part 2 consists of the remaining portion, i.e. the last three quarters of the eighth chapter, and is referred to as the '*tripādī*'.

The *sandhi*-related aphorisms are strewn mainly across Chapters 6 and 8, with one relevant aphorism from Chapter 4 and certain guiding aphorisms that are required for proper understanding and application of the *sandhi* rules, in Chapter 1. The commentary *Kāśikā* explains the work, preserving the original order of the aphorisms. The *Siddhānta-kaumudī*, however, deals with the subject matter of the work category-wise, and devotes five chapters exclusively to *sandhi sūtras*. However, the treatment in this text may be suitable for scholars, but not for beginners. This is because the order of presentation of the *sūtras* in this commentary does not bring clarity to aspects such as precedence rules, and is tuned more towards meaningful modularization. Furthermore, there are *sandhi sūtras* dealt with by the *Siddhānta-kaumudī* in other chapters that deal with case inflections, etc., and not in the *sandhi*-related chapters.

As presented by the authors in [4], the presentation of *sūtras* follows different patterns in the *sapādasaptādhyāyī* and in the *tripādī*. Precedence rules are different in both these parts. Also, exceptions to rules are also presented in different ways. Sometimes exceptions to rules are presented later, while in other cases, they are dealt with earlier. Moreover, an exception to a rule may be present in a completely different chapter itself. For all these reasons, existing sandhi-tutoring books and tools present one or other of the existing schemes and differ only in terms of the examples dealt with or the depth of information presented.

Thus, there is a need for an ontology for *sandhi*-tutoring, which is comprehensive, effective and non-intimidating, and would even help a beginner become thoroughly knowledgeable in the rules of *sandhi* in Sanskrit. It is clear that balancing the requirements of meaningful modularization and clear specification of precedence rules, poses the key challenge to designing such an ontology. Another subsidiary challenge is to present all the information to the learner in a gradual manner so that the learning process is smooth and not intimidating, though terse and complex aphorisms of Pāṇini are being taught.

## 4. The Ontology

For this purpose, *sandhis* were identified as belonging to the following broad categories, rather than those presented in 2.2 above. This categorization is an accepted one prevalent among Sanskrit teachers and teaching material.

1. *Prakṛtibhāva sandhi* – no change is effected in the words
2. Vowel *sandhi* – the $x$ and $y$ involved are both vowels or a combination of vowel and semi-vowel
3. Consonant *sandhi* – at least one of $x$ and $y$ is a consonant
4. *Rutva sandhi* – one of $x$ and $y$ is the special character '*ru*' which is an intermediate output of *sandhi* transformations and is represented in this work by the character '#'. The interpretation of this character is the semi-vowel '*r*'.
5. *Visarga sandhi* – $x$ is a *visarga* or is transformed into a *visarga*.
6. *Anusvāra sandhi* - $x$ is an *anusvāra* or is transformed into an *anusvāra*.

Let $\alpha$, $\beta$, $\gamma$, $\delta$, $\varepsilon$ and $\eta$ be the set of *sūtras* that fall under the above six categories respectively and let $\Omega$ denote the set of all the 104 *sūtras* identified for the purpose of this work. The cardinality of $\Omega$ is 104. Let $M$ be the set of modules of the ontology. Let $\lambda$ denote the ordering function. If $\Theta$ denotes the ontology, then $\Theta = \{M, \omega, \lambda : \omega \in \Omega\}$. $\lambda$ is a one-to-one, onto function $\lambda: \Theta \to \mathbb{N}$, defined as $\lambda(\omega) = n$ where $1 <= n <= 104$.

### 4.1. Structuring the Modules of the Ontology

The core of the ontology developed in this work for *sandhi*-tutoring consists of modules $M_i$, $1 <= i <= 13$, apart from three preliminary modules covering the essential terminology, etc. covered in Section 2 above, as well as general information about the ordering rule followed in the *Āṣṭādhyāyī*, described by the authors in [4]. Each module has been designed to be simple and logically cohesive, with the largest module consisting of 14 *sūtras*. For the purpose of effective modularization, the sets $\beta$, $\gamma$, $\delta$, $\varepsilon$ and $\eta$ were further divided into subsets $\beta_i$, $\gamma_j$, $\delta_k$, $\varepsilon_l$, $\eta_m$, where $1 <= i <= 2$, $1 <= j <= 7$, $1 <= k <= 3$, $1 <= l <= 2$, $1 <= m <= 3$. Clearly, except for the consonant *sandhi sūtras*, which is the largest category having 37 *sūtras* in all, other sets have only a 2-way or 3-way split. This fact contributes to the cohesiveness of the modular design. The general structuring of these core modules is described below.

1. $M_1$: $\alpha$
    a. *Sūtras* defining *pragṛhya* or words that do not allow *sandhi*
    b. *Prakṛtibhāva sandhi sūtras*
    c. *Sūtras* forming exceptions to *pluta* being treated as unchanging

2. $M_2$: $\delta_1$ and $\varepsilon_1$, which include
    a. *Visarga-rutva*
    b. Rules changing *rutva* changing to a vowel
    c. *Lopa* rule

3. $M_3$: $\beta_1$, which includes the following *sandhi* types
    a. *Avaṅādeśa*
    b. *Savarṇadīrgha*
    c. *Pararūpa*
    d. *Vṛddhi*
    e. *Guṇa*
    f. *Pūrvarūpa*

4. $M_4$: $\beta_2$, which includes the following *sandhi* types
    a. *Ayāyāvāvādeśa*
    b. *Yaṇādeśa*

5. $M_5$: $\gamma_1$, which includes the following *sandhi* types
   a. *Tugāgama*
   b. *Lopa* rules
   c. *Jaśtva* (external)
   d. Miscellaneous rules involving a particular word

6. $M_6$: $\delta_2$, $\eta_1$, which include the following *sandhi* types
   a. *Satva*

7. $M_7$: $\gamma_2$, $\delta_3$, which include the following *sandhi* types
   a. Replacement of *rutva*
   b. *Lopa* rules

8. $M_8$: $\gamma_3$, $\eta_2$, which include the following *sandhi* types
   a. *Anusvāra* and related consonant rules

9. $M_9$: $\gamma_4$, which includes the following *sandhi* types
   a. *Dhuḍāgama*
   b. *Tugāgama*
   c. *Ṅamuḍāgama*

10. $M_{10}$: $\varepsilon_2$, which includes the following *sandhi* types
    a. Changes from *visarga* to consonants

11. $M_{11}$: $\gamma_5$, which includes the following *sandhi* types
    a. Internal *sandhi* rule
    b. *Ścutva*
    c. *Ṣṭutva*

12. $M_{12}$: $\gamma_6$, which includes the following *sandhi* types
    a. *Anunāsikā*
    b. Doubling rules
    c. *Jaśtva* (internal)
    d. *Cartva*

13. $M_{13}$: $\gamma_7$, $\eta_3$, which include the following *sandhi* types
    a. *Parasavarṇa*
    b. *Pūrvasavarṇa*
    c. *Chatva*
    d. Some *lopa* rules (internal)

### 4.2. The Ordering Structure in the Ontology

One of the crucial and unique features of this ontology is that the ordering of the *sūtras* is maintained in such a way that only the knowledge gained in previously covered modules are used in later modules. Further, the output of a *sandhi sūtra* with $\lambda = n$, can be provided as input only to *sūtras* with $\lambda > n$. The *sūtras* have been organized in a novel way that is different from what is prevalent in the literature - the *Aṣṭādhyāyī*, the commentaries and teaching material on *sandhis* - ordering, in order to bring clarity while tutoring. The accuracy, strength and effectiveness of this ontology for teaching *sandhis*, have as their bases, this precise ordering coupled with the high degree of cohesiveness in each module.

An example is cited here for the purpose of illustrating the importance of this order and the effectiveness of this modularization. When the words conjoined are *mṛt* (meaning 'mud') and *mayam* (meaning 'made of'), then the rules satisfied in succession in the order specified by the function $\lambda$ developed in this work, is as follows (to the exclusion of the doubling rules):

| Words | $\omega$ (*Aṣṭādhyāyī sūtra*) | *Aṣṭādhyāyī sūtra* number | $\lambda$ | $M_i$ |
|---|---|---|---|---|
| *mṛt* + *mayam* | - | - | - | - |
| *mṛd* + *mayam* | *jhalām jaśo'nte* // | 8.2.39 | 42 | $M_5$ |
| *mṛn* + *mayam* | *yaro'nunāsike'nunāsiko vā* // | 8.4.45 | 86 | $M_{12}$ |

Now *mṛn* + *mayam* would fulfil the requirements of the rule conveyed by the *sūtra*, "*raṣābhyāṁ no ṇaḥ samānapade // 8.4.1 //*", and become transformed into *mṛṇ* + *mayam*. However, the serial number of this *sūtra* in $\lambda$ is 83, and is in $M_{11}$. Therefore, this rule cannot be applied after a rule contained in a later module ($M_{12}$) has been applied. Hence the student who learns using this ontology, arrives at the correct final form *mṛn* + *mayam* and does not wrongly apply rule number 83 thereafter.

### 4.3. Other features of the Ontology

The ontology also contains information regarding whether a *sandhi sūtra* is optional or mandatory, whether it causes an internal transformation within a word or a change between two words, and whether it requires further semantic information for processing. The ontology further includes examples for each type of sandhi along with explanations of the concerned *sūtra*. Examples are carried forward from previous modules to illustrate successive applications of rules. Every module includes a test at the end, to keep track of student progress. The tool developed based on this ontology provides a summary of test scores of each module. The entire tutoring ends with a comprehensive test on all that was taught.

## 5. Conclusion

The ontology implemented in the form of a multimedia *sandhi* tutor that displays Sanskrit in both *Devanāgarī* script as well as using Latin Unicode, was given for testing to three categories of users: Sanskrit scholars ($U_1$), those who have some knowledge of *sandhis* already ($U_2$) and beginners who have an exposure to Sanskrit but have no knowledge of *sandhis* ($U_3$). Feedback on various aspects was solicited from these different categories of users.

The $U_1$ group consisting of four scholars, was asked to individually test the tool rigorously and provide feedback on three criteria in terms of a 5-point scale. The criteria were cohesiveness of the modules, correctness of content and flow, and comprehensiveness of content. The correlation between the first two criteria was found to be 0.71, which is significant. The data indicated that the scholars rated the tool highly on both these criteria, showing that a balance had been struck by the ontology between these two critical criteria. The average rating for comprehensiveness of content was 4.25, which is also high. This validation of the trainer by scholars has served to validate even the methodology fundamentally meant for processing euphonic conjunctions.

Table 1 shows the feedback data got from $U_1$. (Value 5 indicates the highest rating.)

**Table 1:** Feedback data from $U_1$

| Scholar # | Cohesiveness of Modules | Correctness of Content & Flow | Comprehensiveness of Content |
|---|---|---|---|
| 1 | 4 | 5 | 4 |
| 2 | 5 | 5 | 5 |
| 3 | 3 | 4 | 3 |
| 4 | 4 | 4 | 5 |

Table 2 encapsulates the results of the two-tailed $t$-test performed on the feedback got from $U_2$ (consisting of 6 people) and $U_3$ (consisting of 8 people):

**Table 2: Two-tailed t-test on $U_1$ and $U_2$**

| Criterion | $t$ value |
|---|---|
| Ease of use | 2.03 |
| Relevance of examples | 0.99 |
| Usefulness of tests | 1.65 |

Since the tabulated $t_{12}$ at 5% level of significance is 2.18, the conclusion arrived at is that there is no significant difference in the experience had by the two groups with respect to all three criteria. This shows that a novice learner finds the tool useful as well.

Thus, the ontology designed and presented in this work, which is the first of its kind in the literature for comprehensively teaching the rules of euphonic conjunctions of Sanskrit grammar, was validated by various categories of users. This ontology represents a way of ensuring that there is no trade-off between the cohesiveness of modules and the correctness of content and flow, though the nature of the content presented a significant challenge in this context. Furthermore, the ontology also maintains simplicity and clarity of presentation for the student-user.